\documentclass[10pt,twocolumn,letterpaper]{article}

\usepackage{cvpr}              

\usepackage[dvipsnames]{xcolor}

\usepackage{bbding}
\usepackage{tabulary,multirow,xspace}
\usepackage{color}
\usepackage{colortbl}
\usepackage{xcolor}
\definecolor{mygray}{gray}{0.95}
\usepackage{url}            
\usepackage{booktabs}       
\usepackage{amsfonts}       
\usepackage{nicefrac}       
\usepackage{microtype}      
\usepackage{makecell}
\usepackage{adjustbox}
\usepackage{pifont}
\usepackage{float}
\usepackage{stfloats}
\usepackage{wrapfig}
\usepackage{cuted}
\usepackage{mathrsfs}
\usepackage{amsmath}
\usepackage{bm}
\usepackage[cmintegrals]{newtxmath}
\newcommand \footnoteONLYtext[1]
{
	\let \mybackup \thefootnote
	\let \thefootnote \relax
	\footnotetext{#1}
	\let \thefootnote \mybackup
	\let \mybackup \imareallyundefinedcommand
} 
\usepackage[misc]{ifsym} 
\usepackage{ulem} 
\usepackage{arydshln}
\usepackage[accsupp]{axessibility}

\definecolor{cvprblue}{rgb}{0.21,0.49,0.74}
\usepackage[pagebackref,breaklinks,colorlinks,citecolor=cvprblue]{hyperref}

\def\paperID{15338} 
\def\confName{CVPR}
\def\confYear{2024}

\title{Skeleton-in-Context: Unified Skeleton Sequence Modeling with \\ In-Context Learning\vspace{-0.7em}}

\author{Xinshun Wang$^{1,2,*}$, Zhongbin Fang$^{1,*}$, Xia Li$^{3}$, Xiangtai Li$^{4}$, Mengyuan Liu$^{2,~\textrm{\Letter}}$ \\
  \small{$^{1}$Sun Yat-sen University}\\
  \small{$^{2}$National Key Laboratory of General Artificial Intelligence, Peking University, Shenzhen Graduate School} \\
  \small{$^{3}$Department of Computer Science, ETH Zurich}
  \small{$^{4}$S-Lab, Nanyang Technological University}\\
  {\tt\small \{wangxsh36, fangzhb5\}@mail2.sysu.edu.cn, xia.li@inf.ethz.ch, xiangtai94@gmail.com}\\ 
  {\tt\small liumengyuan@pku.edu.cn} \vspace{0.5mm} \\
  {\small\url{https://github.com/fanglaosi/Skeleton-in-Context}\vspace{-2.5mm}}
}

\begin{document}
\maketitle
\begin{abstract}
\vspace{-1em}
\footnoteONLYtext{
*~Equal contribution. \textrm{\Letter}~Corresponding author is Mengyuan Liu.
}
In-context learning provides a new perspective for multi-task modeling for vision and NLP.
Under this setting, the model can perceive tasks from prompts and accomplish them without any extra task-specific head predictions or model fine-tuning.
However, skeleton sequence modeling via in-context learning remains unexplored.
Directly applying existing in-context models from other areas onto skeleton sequences fails due to the similarity between inter-frame and cross-task poses, which makes it exceptionally hard to perceive the task correctly from a subtle context.
To address this challenge, we propose Skeleton-in-Context (SiC), an effective framework for in-context skeleton sequence modeling. 
Our SiC is able to handle multiple skeleton-based tasks simultaneously after a single training process and accomplish each task from context according to the given prompt. 
It can further generalize to new, unseen tasks according to customized prompts.
To facilitate context perception, we additionally propose a task-unified prompt, which adaptively learns tasks of different natures, such as partial joint-level generation, sequence-level prediction, or 2D-to-3D motion prediction. 
We conduct extensive experiments to evaluate the effectiveness of our SiC on multiple tasks, including motion prediction, pose estimation, joint completion, and future pose estimation. We also evaluate its generalization capability on unseen tasks such as motion-in-between. 
These experiments show that our model achieves state-of-the-art multi-task performance and even outperforms single-task methods on certain tasks.
\end{abstract}
\vspace{-2em}
\section{Introduction}
\label{sec:introduction}

\begin{figure}[t]
\centering
\includegraphics[width=0.99\linewidth]{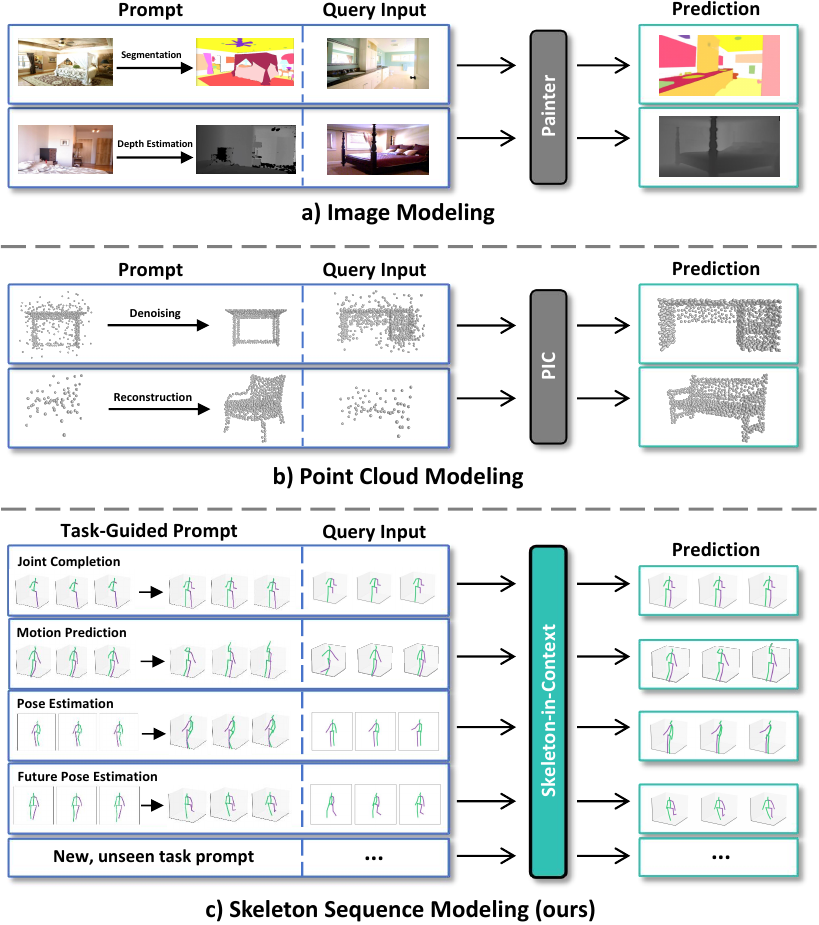}
\centering
\vspace{-0.5em}
\caption{
In-context learning in a) image modeling~\cite{wang2023painter}, b) point cloud modeling~\cite{fang2023pic}, and c) skeleton sequence modeling~(ours).
}
\vspace{-2em}
\label{fig:teaser}
\end{figure}

The idea of multi-task models has attracted much research interest over the years in a wide range of areas~\cite{brown2020language,wang2023painter,fang2023pic,foo2023UPS}, such as computer vision and natural language processing (NLP).
To build such models, most previous works focus on deploying task-specific heads on a shared backbone and learning them through pretraining and fine-tuning. Models with such task-specific architectures are not learned in an end-to-end manner and are hard to generalize to new, unseen tasks and data.
Most existing methods for skeleton-based human-centric tasks rely on task-specific designs and/or non-end-to-end frameworks to accomplish multiple tasks.
Several works~\cite{cai2021unified,ma2022diffsion} learn unified human motion representations. 
In particular, MotionBERT~\cite{zhu2023motionbert} adopts two-stage training, including pre-training on the 2D-3D lifting task and fine-tuning on each task. 
UniHCP~\cite{ci2023unihcp} connects a well-designed shared backbone network to specific task heads according to different tasks. 
Meanwhile, UPS~\cite{foo2023UPS} unifies action labels and motion sequences into text-based language sequences and fine-tunes them on each task for better performance. 
These models can have a limited scope of tasks they are capable of handling and lack flexibility.
This limits their capability to generalize to unseen tasks and data.
To overcome these challenges, in-context learning represents a new trend to build generalist models, which was originally proposed in NLP~\cite{brown2020language,rubin2021learning,radford2021learning} and then introduced into other areas such as images and point clouds.
Some generalist models~\cite{bar2022visualprompt,wang2023painter,wang2023seggpt,zhang2023goodexample} utilize the Masked Image Modeling (MIM) framework to exhibit multi-task capability via in-context learning in image processing, as shown in Fig.~\ref{fig:teaser}(a). Similarly, in Fig.~\ref{fig:teaser}(b), Point-In-Context~\cite{fang2023pic} proposes a key joint sampling module and introduces in-context learning into point cloud processing.
Specifically, in-context models learn patterns of provided input-output pairs, called in-context examples or prompts, and then perform the desired task according to the prompt.
An in-context model can accomplish diverse tasks end-to-end without any task-specific heads or fine-tuning. It also has excellent generalization capability.
Directly adapting existing in-context models from other areas onto skeleton sequences fails because they largely focus on static objects (images or point clouds) and thus cannot capture spatial-temporal dependencies in skeleton sequences.
To the best of our knowledge, in-context learning remains unexplored in 3D skeleton sequence modeling.

In this paper, we propose Skeleton-in-Context (SiC), a novel framework for skeleton sequence modeling, which is able to process multiple tasks simultaneously after a single training process without any task-specific designs, as Fig.~\ref{fig:teaser}(c) shows.
Specifically, SiC perceives tasks from context, i.e., the prompt from a desired task provided as additional input, and then accomplishes the task for the query.
SiC employs two types of prompts to facilitate context perception: the task-guided prompt~(TGP) and the task-unified prompt~(TUP). 
TGP provides a task template for the model, giving the model the ability to make analogies. This is the key to the model's capability to generalize to new tasks, as we can guide the model to do unseen tasks by customizing specific prompts. 
TUP can be implemented in two ways: static and dynamic. 
The static TUP leverages prior knowledge shared among tasks, while the dynamic TUP adaptively learns to perceive task context.
Skeleton-in-Context provides a new perspective for training a skeleton-based multi-task in-context model. 
Since there is no existing benchmark for measuring such a model, we establish a new in-context benchmark based on four tasks, including three skeleton-based tasks: motion prediction, pose estimation, and joint completion, and a new task: future pose estimation.

Our main contributions are as follows:
1) We introduce a novel framework, Skeleton-in-Context~(SiC), designed to handle multiple skeleton-based tasks simultaneously. Our approach learns the potential pattern of the given prompt and performs it on the query sample.
2) We establish a new 3D human in-context learning benchmark comprised of multiple skeleton-based tasks, such as motion prediction, pose estimation, joint completion, and future motion estimation. We further benchmark several representative in-context learning baselines for reference.
3) We conduct extensive experiments to evaluate our SiC. SiC achieves state-of-the-art results in multi-tasking and even outperforms task-specific models on certain tasks. Moreover, SiC also generalizes well to unseen tasks such as motion in-between. 
\section{Related Work}
\label{sec:related_work}

\noindent
\textbf{Skeleton-Based Motion Analysis.}
Many of the skeleton-based human-centric tasks have shared similar development patterns in terms of the methods used to address them.
On human motion prediction, early successes are achieved with RNNs and CNNs~\cite{martinez2017human,li2018convolutional}. Nowadays, the task is dominated by Graph Convolutional Networks (GCNs).
LTD~\cite{mao2019learning} considers each joint coordinate as a node in a graph on which to employ learnable spatial graph convolutions.
Follow-up works~\cite{mao2020history,cui2020learning,wang2024gcnext,wang2024dynamic} focus on designing better graph convolution networks.
Different from motion prediction, the goal of motion synthesis is to complete the human motion sequence.
To complete the missing parts of human motion, many works adopt convolutional models~\cite{yan2019convolutional,kaufmann2020convolutional,henter2020moglow}or generative adversarial networks~\cite{zhou2020generative,cai2018deep,harvey2020robust,hernandez2019human}.
One category of 3D pose estimation~\cite{gong2023diffpose,liu2019feature,liu2020comprehensive} is to lift the estimated 2D pose to 3D, which often employs spatial-temporal convolution~\cite{martinez2017simple,pavllo20193d,cheng20203d,cai2019exploiting,ci2019optimizing,wang2020motion,xu2021graph} or transformer-based~\cite{zhang2022mixste,zheng20213dpose,shan2022p,zhao2022graformer,zheng20213dposetrans,vg4dICRA24} methods.
However, these approaches focus on designing customized architectures specific to a task, which are limited in range of applications. To this end, we propose to unify the input and output formats of multiple tasks and utilize a single model to process multiple tasks simultaneously via in-context learning.

\noindent
\textbf{In-Context Learning.}
The emergence of in-context learning provides a new paradigm for achieving multi-tasking, which originated from Natural Language Processing (NLP)~\cite{rubin2021learning,radford2021learning,devlin2018bert,sarzynska2021detecting,alayrac2022flamingo}. It endows models with the capability to handle multiple tasks and even unseen tasks by incorporating domain-specific input-target pairs, known as prompts, into the query sample~\cite{brown2020language}. Following NLP, in-context learning is extended to computer vision in some work, such as 2D images~\cite{wang2023painter,wang2023seggpt,sun2023exploring,zhang2023goodexample,wang2023promptdiffusion,wang2024refldmseg} and 3D point clouds~\cite{wu2023towards,balavzevic2023towards,fang2023pic}. Specifically, Visual Prompt~\cite{bar2022visualprompt}, Painter~\cite{wang2023painter}, and Point-In-Context~\cite{fang2023pic} are pioneers of in-context methods in computer vision. They verify the effectiveness and flexibility of in-context learning in vision-based tasks, and most of them adopt pure-transformer-based approaches~\cite{he2022mae,bao2021beit,li2023transformer,wu2023open}.
Meanwhile, several approaches deeply explore Prompt Engineering~\cite{sun2023exploring,zhang2023goodexample,balavzevic2023towards} and awe-inspiring generalization~\cite{wang2023seggpt,fang2023pic,wu2023towards} in in-context learning.
Recently, LVM~\cite{bai2023sequential} unifies most vision tasks and data via joint co-training on one visual in-context model. Compared to the works focused on \textit{static} objects~(images and point clouds), our work mainly delves into modeling in-context \textit{dynamic} skeleton sequences.

\noindent
\textbf{Unified Skeleton-Centric Models.}
A generalist model that can excel in handling multiple tasks is a profound step toward general artificial intelligence~\cite{wang2023painter,radford2021learning}. 
Previous works~\cite{kalayeh2018SPReID,khamis2015joint,su2017multi,tian2015pedestrian,lin2018multigrained,xu2022fashionformer,luvizon2018poseaction,yao2012coupled} give the effort to explore co-training several human-centric tasks. 
Meanwhile, some works~\cite{liang2018lookintoperson,nie2018mutual} show great progress in simultaneously predicting parsing and pose. 
Recent research~\cite{cai2021unified,ma2022diffsion} demonstrates the effectiveness of modeling unified human motion sequences. 
In particular, UniHCP~\cite{ci2023unihcp} designs specific task heads according to different 2D human-centric tasks. 
Motion-BERT~\cite{zhu2023motionbert} is pre-trained in the 2D-3D lifting task and fine-tuned on each task. 
UPS~\cite{foo2023UPS} unifies heterogeneous human behavior understanding tasks via language sequences and cycle training for each task in order.
Nevertheless, these works are limited by the dilemma of complex training stages or task heads, which introduce extra costs. We integrate the output format of different tasks in skeleton sequence format and train the model just once.

\section{In-Context Skeleton Sequence Modeling}
\label{sec:benchmark}

\subsection{Formulation}
\label{sec:incontext_modeling}

Since the skeleton sequence data in skeleton-based tasks are sequential by nature, we treat the skeleton at each frame as a token, which can be processed by transformer~\cite{vaswani2017attention}.
Following previous works~\cite{wang2023painter,bar2022visualprompt,fang2023pic}, we curate an in-context skeleton sequence paradigm. Given an input-target skeleton sequence pair as a task-guided prompt $G^{k}_{i}=\left \{I^{k}_{i}, T^{k}_{i} \right \}$ and a query input $Q^{k}_{j}$, where $I^{k}_{i}, T^{k}_{i},Q^{k}_{j}\in \mathbb{R}^{F\times J\times 3}$ each have $F$ frames and $J$ joints, the model is expected to perceive the context from the prompt, i.e., which task it represents, and perform the corresponding operations on the query in analogy to the prompt. This process can be formulated as:
\begin{equation}
    y=f(G^{k}_{i},Q^{k}_{j}),
\end{equation}
where $k$ represents the task index, while $i$ and $j$ indicate different example indexes.
Thus, we can determine what task will be performed on the query sample by selecting the task-specific prompt from the corresponding training set. 

\begin{figure}[t]
\centering
\includegraphics[width=0.99\linewidth]{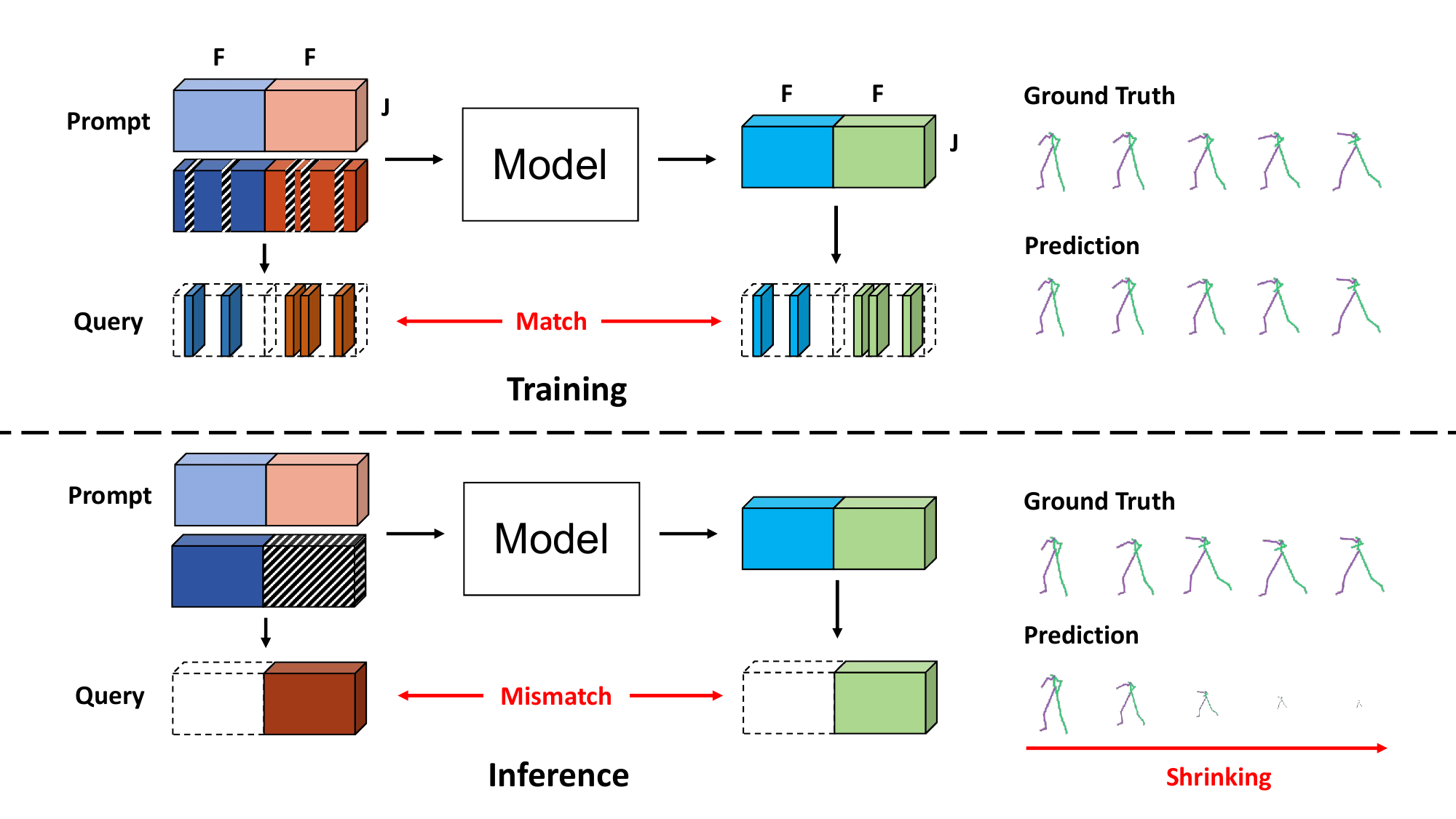}
\centering
\vspace{-0.5em}
\caption{
\textit{Top:} Training with MIM-style framework~\cite{he2022mae,pang2022pointmae} used in previous works~\cite{wang2023painter,bar2022visualprompt,fang2023pic}. The model is able to reconstruct the masked frames well.
\textit{Bottom:} During inference, the reconstructed sequence gradually shrinks the skeleton as time goes by because the model only learns frame interpolation during training. Once the model loses subsequent reference frames when interpolating frames, the generated sequence will tend to shrink.
}
\vspace{-1.5em}
\label{fig:overfitting}
\end{figure}

\begin{figure*}[t]
\centering
\includegraphics[width=0.95\linewidth]{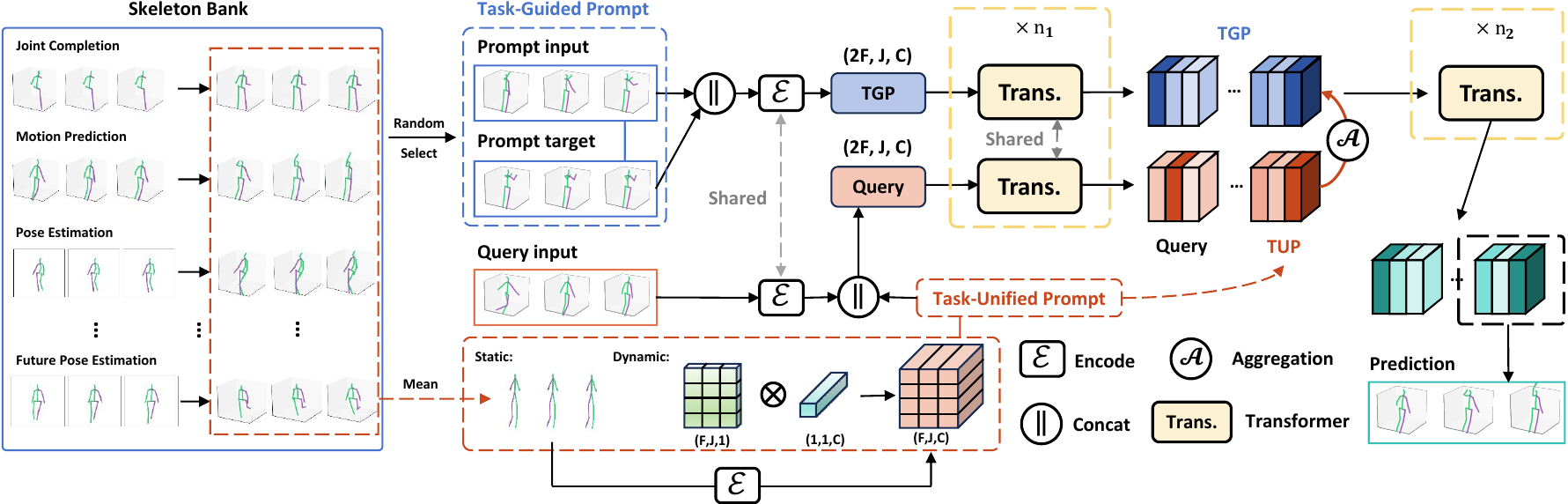}
\centering
\vspace{-0.5em}
\caption{Overall framework of our Skeleton-in-Context. Specifically, we establish a skeleton bank by integrating training sets under different tasks, which contain a large amount of input-target pairs performing different tasks. Next, we randomly select a sample pair as the task-guided prompt (TGP) and a query input from the skeleton bank, undergo encoding and concatenating, respectively, and then input them into the transformer in parallel. In particular, during this process, the query input and task-unified prompt (TUP) are combined to form a new query. After iterating $n_{1}$ times, the TGP and query are aggregated through $\mathcal{A}\left( \cdot\right)$ and then input into the transformer for $n_{2}$ iterations. Lastly, the second half of the model output is used as our prediction.
}
\vspace{-0.5em}
\label{fig:framework}
\end{figure*}

\subsection{Task Definitions}

To measure multitasking capability, we establish a new in-context benchmark from common skeleton-based datasets, including H3.6M~\cite{ionescu2013h36m}, AMASS~\cite{mahmood2019amass}, and 3DPW~\cite{von2018_3dpw}. In this benchmark, we select three common tasks: motion prediction, pose estimation, and joint completion, and a new task: future pose estimation. Consequently, we conduct extensive experiments on a large and diverse dataset collection, which consists of 176K skeleton sequence samples.

\noindent
\textbf{Motion Prediction.}
Motion prediction~\cite{mao2019learning}is to predict future motion based on historical motion.
There is continuity between the input and output sequences, which constitute a complete motion. Thus, the input $I$ and target $T$ are complete skeleton sequences with shapes of $(F,J,3)$.

\noindent
\textbf{Pose Estimation.}
Pose estimation~\cite{gong2023diffpose,liu2019feature,liu2020comprehensive} is to lift 2D poses to 3D. The model takes the detected 2D skeleton sequence $I\in \mathbb{R}^{F\times J\times 2}$, which contain noise, as input, and estimates the accurate 3D skeleton sequence $T\in \mathbb{R}^{F\times J\times 3}$.
Note that we expand the 2D input to 3D by filling the 3rd dimension with $0$, in order to unify the data format with other tasks. Therefore, the input skeleton sequence becomes the same shape $(F,J,3)$ as the target sequence.

\noindent
\textbf{Future Pose Estimation.}
Based on prediction and estimation, we propose a new task: future pose estimation, which is to predict future 3D motion based on history 2D poses. As 2D images are easier to obtain than 3D data, this task unlocks a host of new real-world applications.
Given a historical 2D skeleton sequence $I\in \mathbb{R}^{F\times J\times 3}$~(one dimension with $0$), future pose estimation expects the model to output its future 3D skeleton sequence $T\in \mathbb{R}^{F\times J\times 3}$. To our knowledge, no work has ever proposed future pose estimation.

\noindent
\textbf{Joint Completion.}
Joint completion is to recover skeleton sequences that have missing joints.
This task can reflect cases where joints may be missing or displaced when collected with Lidar cameras~\cite{shahroudy2016ntu60,liu2019ntu120}.
We set two levels to verify the ability to complete joints, which are 40$\%$ and 60$\%$ of the joints missing in each frame. Also, the coordinates of missing joints are replaced with 0, with the goal of unifying the input skeleton sequence format with $(F,J,3)$.

\section{Skeleton-in-Context}
\label{sec:method}

In-context learning on other data modalities such as images and point clouds~\cite{bar2022visualprompt,wang2023painter,fang2023pic} largely employ Mask Image Modeling~(MIM)~\cite{he2022mae} to reconstruct the randomly masked elements. During inference, only the target elements are masked. On skeleton sequences, MIM fails due to spatial-temporal complexity and inter-pose
similarity, which makes it outstandingly hard to perceive the task correctly from a subtle context.
In our Skeleton-in Context, we employ a new framework to address this challenge.

\subsection{Skeleton Sequence Prompt}

\noindent
\textbf{Task-Guided Prompt.}
In in-context learning, a prompt is needed for the model to learn the underlying task pattern from context.
As mentioned in Sec.~\ref{sec:incontext_modeling}, we propose the Task-Guided Prompt~(TGP) $G^{k}_{i}=\left \{I^{k}_{i}, T^{k}_{i} \right \}$, which comprises of an example skeleton pair performing a specific task $k$. In this way, our proposed model is able to perform a variety of tasks according to the provided prompts that could be seen or even unseen in the training set. It is worth noting that our model can perform unseen tasks and transfer to datasets that are different styles from the training dataset. It will be elaborated in Sec.~\ref{sec:mian_results}.

\noindent
\textbf{Avoiding Over-Fitting.}
Previous works have made great progress in unifying different tasks via the MIM-style framework~\cite{bar2022visualprompt,wang2023painter,fang2023pic}.
Compared with static objects in 2D and 3D, such as images and point clouds, the obstacle to in-context learning of skeleton sequences lies in the wondrous similarity of each frame, so directly using the random masking strategy~\cite{wang2023painter,bar2022visualprompt,fang2023pic} will cause the model to fall into over-fitting, leading it to only learn simple interpolation frames rather than capturing meaningful motion sequence representations during the training stage.
As shown in Fig.~\ref{fig:overfitting}, when we directly utilize the MPM-style~\cite{he2022mae,pang2022pointmae} framework to train our model, although it can perfectly reconstruct the masked frame during training, the experimental results reveal that it outputs a devastating result during inference. It is obvious that the output skeleton sequence gradually shrinks as the length increases because the next frame of the current frame is 0, and interpolating with 0 will lead to shrinking. Such a phenomenon underscores the challenges of learning contextual information about dynamic skeleton sequences.

\begin{table*}[t]
\small
\centering
\caption{Comparison of task-specific models, multi-task models, and in-context models on four human skeleton tasks. For Motion Prediction~(MP.), Joint Completion~(JC.), and Future Pose Estimation~(FPE.), we report Mean Per Joint Position Error~(mm)~\cite{ionescu2013h36m}. For Pose Estimation~(PE.), we additionally report another indicator, Normalized Mean Per Joint Position Error~(N-MPJPE)~\cite{zhu2023motionbert}. For joint completion, we report two settings: 40\% and 60\% of the joints missing in each frame. $\dagger$ represents the multi-task-capable models re-structured from task-specific models. $\ddagger$ represents the multi-task-capable models re-structured from multi-stage models. Static-SiC and Dynamic-SiC represent the static and dynamic versions of TUP, respectively.}
\vspace{-0.5em}
\setlength\tabcolsep{0.5mm}
\scalebox{0.9}{
\begin{tabular}{lccccccccccccccc}
\hline
\multicolumn{1}{l|}{\multirow{3}{*}{Methods}} & \multicolumn{1}{c|}{\multirow{3}{*}{Venues}} & \multicolumn{6}{c|}{MP. (AMASS)}     & \multicolumn{2}{c|}{PE. (H3.6M)} & \multicolumn{3}{c|}{JC. (3DPW)} & \multicolumn{3}{c}{FPE. (H3.6M)} \\
\multicolumn{1}{c|}{}  & \multicolumn{1}{c|}{} & \multicolumn{6}{c|}{MPJPE $\downarrow$}     & MPJPE$\downarrow$ & \multicolumn{1}{c|}{N-MPJPE$\downarrow$}   & \multicolumn{3}{c|}{MPJPE $\downarrow$} & \multicolumn{3}{c}{MPJPE $\downarrow$} \\
\multicolumn{1}{c|}{}  & \multicolumn{1}{c|}{}  & 80ms & 160ms & 200ms & 320ms & 400ms & \multicolumn{1}{c|}{Avg.} & Avg. & \multicolumn{1}{c|}{Avg.}        & 40\%   & 60\%   & \multicolumn{1}{c|}{Avg.}   & 200ms  & 300ms & Avg.      \\ \hline
\multicolumn{16}{c}{Task-Specific Model: one architecture for one task}   \\ \hline
\rowcolor{mygray}
\multicolumn{1}{l|}{siMLPe~\cite{guo2023back}}  & \multicolumn{1}{c|}{WACV'23}    & 12.1    & 21.6     & 25.4   & 34.4 & 39.9 & \multicolumn{1}{c|}{26.7}  & 58.9   & \multicolumn{1}{c|}{49.7}    & 48.4      & 58.5      & \multicolumn{1}{c|}{53.5}      & 75.4    & 77.6         & 76.5     \\
\rowcolor{mygray}
\multicolumn{1}{l|}{EqMotion~\cite{xu2023eqmotion}}   & \multicolumn{1}{c|}{CVPR'23}        & 14.4    & 26.6     & 31.7  & 44.0 & 50.0   & \multicolumn{1}{c|}{33.3}  & 63.2   & \multicolumn{1}{c|}{59.3}   & 40.4      & 49.7      & \multicolumn{1}{c|}{45.1}      & 69.5     & 70.8  & 70.1    \\
\rowcolor{mygray}
\multicolumn{1}{l|}{STCFormer~\cite{tang2023stcformer}}   & \multicolumn{1}{c|}{CVPR'23}        & 19.5    & 25.1     & 28.2  & 35.8 & 41.5   & \multicolumn{1}{c|}{28.7}  & 54.6   & \multicolumn{1}{c|}{45.2}   & 31.2      & 46.7      & \multicolumn{1}{c|}{38.9}      & 60.9     & 81.8  & 71.3    \\
\rowcolor{mygray}
\multicolumn{1}{l|}{GLA-GCN~\cite{yu2023glagcn}}   & \multicolumn{1}{c|}{ICCV'23}        & 19.5    & 25.1    & 28.2  & 35.3 & 42.0   & \multicolumn{1}{c|}{30.0}  & 49.6   & \multicolumn{1}{c|}{40.4}   & 62.2      &  72.5     & \multicolumn{1}{c|}{67.4}      & 49.5     & 51.4  & 50.5    \\
\rowcolor{mygray}
\multicolumn{1}{l|}{MotionBERT~\cite{zhu2023motionbert}}   & \multicolumn{1}{c|}{ICCV'23}        & 12.9    & 21.4    & 25.5  & 36.3 & 42.3   & \multicolumn{1}{c|}{26.7}  & 49.4   & \multicolumn{1}{c|}{41.1}   & 29.3      &  44.3     & \multicolumn{1}{c|}{36.8}      & 75.5     & 100.3  & 87.9    \\ \hline
\multicolumn{16}{c}{Multi-Task Model: one architecture (w. multiple task-specific heads) for multiple tasks}    \\ \hline
\multicolumn{1}{l|}{siMLPe$^{\dagger}$~\cite{martinez2017simple}}                  & \multicolumn{1}{c|}{WACV'23}             & 19.0          & 30.0    & 35.3     & 46.8  & 52.4   & \multicolumn{1}{c|}{36.7}  & 74.6   & \multicolumn{1}{c|}{60.0}      & 59.0      & 67.4      & \multicolumn{1}{c|}{63.2}      & 64.4     & 68.4       & 66.4                 \\
\multicolumn{1}{l|}{EqMotion$^{\dagger}$~\cite{xu2023eqmotion}}      & \multicolumn{1}{c|}{CVPR'23} & 14.0  & 25.3    & 30.2   & 40.5  & 46.6     & \multicolumn{1}{c|}{31.3}  & 163.3   & \multicolumn{1}{c|}{110.1}         & 38.9      & 56.0      & \multicolumn{1}{c|}{47.5}      & 88.1       & 89.1   & 88.6                 \\ 
\multicolumn{1}{l|}{STCFormer$^{\dagger}$~\cite{tang2023stcformer}}      & \multicolumn{1}{c|}{CVPR'23} & 17.4  & 28.0    & 31.5   & 43.0  & 40.4     & \multicolumn{1}{c|}{33.9}  & 55.6   & \multicolumn{1}{c|}{45.6}         & 33.6      & 50.6      & \multicolumn{1}{c|}{42.1}      & 62.2      & 71.7   & 67.0                 \\ 
\multicolumn{1}{l|}{GLA-GCN$^{\dagger}$~\cite{yu2023glagcn}}   & \multicolumn{1}{c|}{ICCV'23}        & 35.3    & 39.8     & 42.3 & 49.2 & 56.1   & \multicolumn{1}{c|}{44.5}  & 78.6   & \multicolumn{1}{c|}{62.9}   & 74.1      & 81.9      & \multicolumn{1}{c|}{78.0}      & 60.6     & 68.1  & 64.4    \\
\multicolumn{1}{l|}{MotionBERT$^{\ddagger}$ ~\cite{zhu2023motionbert}}                  & \multicolumn{1}{c|}{ICCV'23}             & 14.3         & 22.8    & 26.7     & 36.4  & 41.7   & \multicolumn{1}{c|}{28.4}  & 55.8   & \multicolumn{1}{c|}{45.7}      & 31.3      & 46.7      & \multicolumn{1}{c|}{39.0}      & 75.5     & 96.8       & 86.2                 \\\hline
\multicolumn{16}{c}{In-Context Model: one task-agnostic architecture for multiple tasks}    \\ \hline
\multicolumn{1}{l|}{Copy~\cite{bar2022visualprompt,fang2023pic}}   & \multicolumn{1}{c|}{-}  & 134.7  & 133.7    & 133.3     & 132.1  & 131.4   & \multicolumn{1}{c|}{133.0}   & 351.0  & \multicolumn{1}{c|}{171.6}    & 196.5      & 195.5      & \multicolumn{1}{c|}{196.0}      & 316.8        & 316.6     & 316.7     \\
\multicolumn{1}{l|}{PointMAE~\cite{pang2022pointmae}}   & \multicolumn{1}{c|}{ECCV'22}  & 19.1  & 31.2    & 36.6     & 54.9  & 67.7   & \multicolumn{1}{c|}{41.9}   & 144.3  & \multicolumn{1}{c|}{96.41}    & 67.6      & 79.0      & \multicolumn{1}{c|}{73.3}      & 139.9        & 142.9     & 141.4                 \\
\multicolumn{1}{l|}{PIC~\cite{fang2023pic}}   & \multicolumn{1}{c|}{NeurIPS'23}     & 25.0  & 40.2    & 46.8  & 67.2   & 79.4     & \multicolumn{1}{c|}{51.7}  & 181.6   & \multicolumn{1}{c|}{118.8}       & 85.7      & 96.4      & \multicolumn{1}{c|}{91.1}      & 173.1        & 176.5 & 174.8                 \\ \hline
\multicolumn{1}{l|}{Static-SiC~(ours)}      & \multicolumn{1}{c|}{-}    & \textbf{\textcolor{blue}{11.0}}   & 19.0     & 22.7   & 32.6 & 37.8 & \multicolumn{1}{c|}{24.6}   & \textbf{\textcolor{blue}{51.6}}  & \multicolumn{1}{c|}{\textbf{\textcolor{blue}{41.8}}}     & 29.8      & 44.7      & \multicolumn{1}{c|}{37.2}      & 59.8    & 67.2    & 63.5           \\ 
\multicolumn{1}{l|}{Dynamic-SiC~(ours)}      & \multicolumn{1}{c|}{-}    & \textbf{\textcolor{blue}{11.0}}   & \textbf{\textcolor{blue}{18.9}}     & \textbf{\textcolor{blue}{22.6}}   & \textbf{\textcolor{blue}{32.2}} & \textbf{\textcolor{blue}{37.4}} & \multicolumn{1}{c|}{\textbf{\textcolor{blue}{24.4}}}   & 51.8  & \multicolumn{1}{c|}{42.5}     & \textbf{\textcolor{blue}{29.5}}      & \textbf{\textcolor{blue}{44.0}}      & \multicolumn{1}{c|}{\textbf{\textcolor{blue}{36.8}}}      & \textbf{\textcolor{blue}{59.2}}    & \textbf{\textcolor{blue}{66.5}}    & \textbf{\textcolor{blue}{62.9}}       \\   
\hline
\end{tabular}
}
\vspace{-1.5em}
\label{tab:main_results}
\end{table*}

\noindent
\textbf{Task-Unified Prompt.}
With the Task-Guided Prompt, which exhibits unique capabilities specific to in-domain or out-of-domain tasks, mitigating the gap between different tasks presents a new challenge. 
Meanwhile, as discussed above, a straight extension of previous works will lead to a simple model that can only perform the interpolation task based on the given skeleton sequence. 
To tackle these challenges, we introduce the Task-Unified Prompt~(TUP), a well-designed and task-agnostic module, which can adaptively learn to incorporate multiple tasks into the framework of in-context learning, i.e., a one-off end-to-end framework for all tasks without having to deploy task-specific heads or fine-tuning.
We introduce two versions of TUP:

$Version~1:$ Static-TUP with static pseudo pose:
\begin{equation}
    U^{S}=\mathcal{E}\left ( \frac{1}{M}\displaystyle\sum_{m=1}^{M}\displaystyle\sum_{k=1}^{K}\left ( T_{m}^{k}\right ) \in {\mathbb{R}}^{F\times J\times 1}\right ),
\end{equation}
where $\mathcal{E} \left( \cdot\right )$ is a embedding layer. $M$ and $K$ indicate the total number of training samples and tasks, respectively.

$Version~2:$ Dynamic-TUP with dynamic pseudo pose:
\begin{equation}
    U^{D}=W_{1} W_{2},
\end{equation}
where $W_{1}\in {\mathbb{R}}^{F\times J\times 1}$ and $W_{2} \in {\mathbb{R}}^{1\times 1\times C}$.

We denote the SiC-based on Static-/Dynamic-TUP is referred to as Static-/Dynamic-SiC, respectively.
The proposed TUP facilitates the modeling of distribution differences among tasks and datasets,
so the model focuses more on learning unified representations shared across tasks.

\subsection{Model Architecture}
\label{model_architecture}
As shown in Fig.~\ref{fig:framework}, we build a skeleton bank, a collection of publicly available datasets including H3.6M, AMASS and 3DPW, covering common skeleton analysis tasks, containing input-target pairs formed as sequences of equal length, where prompts are randomly selected. Given a prompt pair $G^{k}_{i}=\left \{I^{k}_{i}, T^{k}_{i} \right \}$ and a query sample $Q^{k}_{j}$, we first map them to the C-dimensional feature space, and then splice $I$ and $T$, $Q$ and $U$ respectively. The former represents our proposed Task-Guided Prompt, and the latter is the objective sequence of our focused analysis. The process can be formulated as:
\begin{equation}
    \widetilde{G} = \mathcal{E}\left ( G^{k}_{i}\right )  \in \mathbb{R}^{2F\times J\times C},
\end{equation}
\begin{equation}
    \widetilde{Q} = \left ( \mathcal{E}\left ( Q_{j}^{k}\right )\parallel U \right ) \in \mathbb{R}^{2F\times J\times C},
\end{equation}
where $\mathcal{E}\left ( \cdot\right )$ is a linear transformation and $\parallel$ is concatenating operator. In essence, $U$ represents the prior knowledge of the target skeleton sequence and is also an adapter that adapts to different tasks and different distributions of data for learning.

\begin{figure*}[t]
\centering
\includegraphics[width=0.90\linewidth]{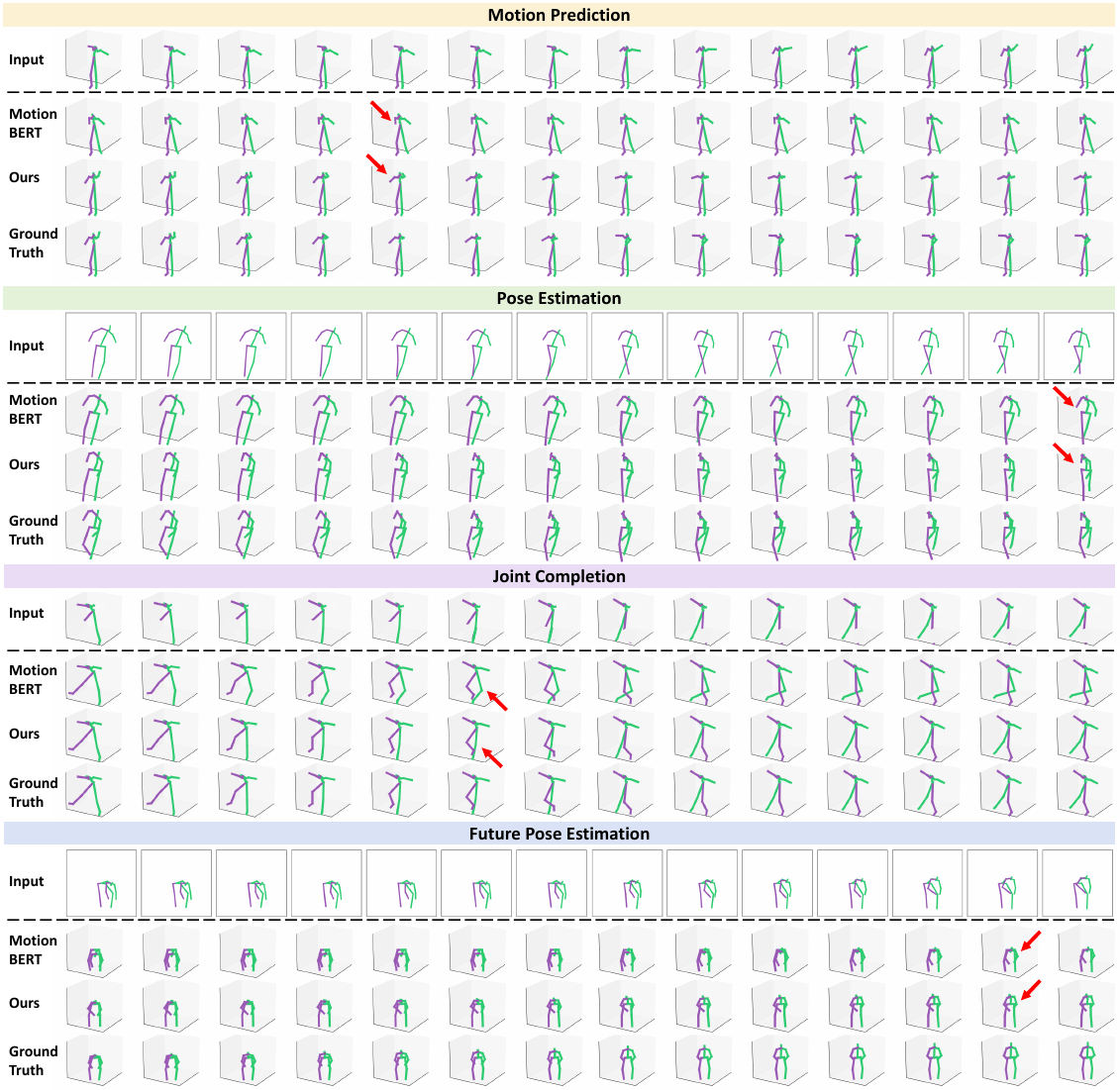}
\centering
\caption{The comparison of visual results between our Skeleton-in-Context and recent SoTA method MotionBERT~\cite{zhu2023motionbert} on four tasks. Our method generates more accurate poses than MotionBERT, as can be seen from the spots with red arrows (\bm{\textcolor{red}{$\searrow$}}) pointed to them.
}
\vspace{-1.5em}
\label{fig:visualization}
\end{figure*}

Drawing inspirations from \cite{zhu2023motionbert}, we connect spatial attention blocks and temporal attention blocks together in an alternating combination manner to form a two-stream transformer $\mathcal{M}\left( \cdot\right)$ dedicated to capturing the dynamic characteristics of motion, which is implemented as:
\begin{equation}
    \mathcal{M}\left( \mathrm{X}\right)
    =\alpha^{i} \mathcal{T}_{1}^{i}\left ( \mathcal{S}_{1}^{i}\left ( \mathrm{X}^{i-1}\right )\right ) + \beta^{i} \mathcal{S}_{2}^{i}\left ( \mathcal{T}_{2}^{i}\left ( \mathrm{X}^{i-1}\right )\right ),
\end{equation}
where $\alpha$ and $\beta$ are balanced parameters determined adaptively. $\mathcal{S}_{1}^{i}$ and $\mathcal{T}_{1}^{i}$ are implemented using spatial self-attention mechanism and temporal self-attention mechanism, respectively. Additionally, $i=1,...,N$.

Then, $\mathcal{M}\left( \cdot\right)$ takes $\widetilde{G}$ and $\widetilde{Q}$ as input and parallel iterates over them $n_{1}$ times respectively. Then, $\widetilde{G}^{n1}$ and $\widetilde{Q}^{n1}$ are integrated via aggregation function $\mathcal{A}\left( \cdot\right)$:
\begin{equation}
    W^{n1}=\mathcal{A}\left( \widetilde{G}^{n1}, \widetilde{Q}^{n1}\right),
\end{equation}
we consider $\mathcal{A}\left( \cdot\right)$ to be a simple weighted summation function. The final output is obtained after iterating $W$ $n_{2}$ times in $\mathcal{M}\left( \cdot\right)$, where $n_1+n_2=N$. Note that we take the second half of the output as the prediction.

\subsection{Training and Inference}

\noindent
\textbf{Training.}
We train our model in our proposed skeleton-centric in-context dataset mentioned in Sec.~\ref{sec:benchmark}. In this training way, our Skeleton-in-Context can perform multiple tasks after one training. Since our output format is skeleton-like, following previous work~\cite{zhu2023motionbert}, we adopt the 3D joint-level reconstruction loss:
\vspace{-1em}
\begin{equation}
Loss=\displaystyle\sum^{F}_{f=1}\displaystyle\sum^{J}_{j=1}\left \|{P}_{f,j}-{G}_{f,j} \right \|_{2},
\end{equation}
\vspace{-1em}

\noindent
\textbf{Inference.}
After a training phase involving all tasks, we get an in-context model that focuses on dynamic skeleton sequences. It can perform multiple tasks at the same time according to the provided task-dependent prompt without any further parameter updates.
\section{Experiments}
\label{sec:experiments}

\noindent
\textbf{Implementation Details.}
We implement the proposed Skeleton-in-Context model with the number of layers $N$ = 5, number of attention heads $H$ = 8, and hidden feature dimension $C$ = 256. For each prompt input/target and query input/target, the sequence length is $F$ = 16.
%
%
All tasks are unified into a one-off, end-to-end training process without any task-specific designs. The training takes about 6 hours for 120 epochs on 4 NVIDIA GeForce RTX 4090 GPUs.
%

\begin{table*}[t]
\small
\centering
\setlength\tabcolsep{2.5mm}
\vspace{-1em}
\caption{Comparison results between task-specific models and our proposed models on generalization testing experiments on different datasets. We report Mean Per Joint Position Error~(mm) for all tasks.}
\vspace{-0.5em}
\scalebox{0.95}{
\begin{tabular}{l|cccccc|cc|ccc}
\hline
\multirow{2}{*}{Methods} & \multicolumn{6}{c|}{MP. (3DPW)}      & \multicolumn{2}{c|}{PE.}       & \multicolumn{3}{c}{JC. (H3.6M)} \\
                         & 40ms & 80ms & 120ms & 160ms & 200ms & Avg. & AMASS  &  3DPW  & 40\%   & 60\%   & Avg.   \\ \hline
\multicolumn{12}{c}{Task-Specific Model: one architecture for one task}   \\ \hline
\rowcolor{mygray}
EqMotion~\cite{xu2023eqmotion}     & 15.6 & 35.6 & 52.1  & 67.9  & 79.5  & 50.1 & 51.8    & 156.6     & 193.6      & 194.0      & 193.8      \\
\rowcolor{mygray}
STCFormer~\cite{tang2023stcformer} & 18.3    & 39.3    & 58.9     & 75.0     & 88.6     & 56.0    & 45.0 & 141.2 & 177.2  & 178.4  & 177.8  \\
\rowcolor{mygray}
siMLPe~\cite{martinez2017simple}   & 14.9 & 26.2 & 44.0  & 56.6  & 68.8  & 42.1 & 48.8    & 150.4    & 236.3  & 257.8  & 247.1  \\ \hline
\multicolumn{12}{c}{In-Context Model: one task-agnostic architecture for multiple tasks}    \\ \hline
Static-SiC~(ours)                  & 7.9  & 21.5 & 37.7  & 54.0  & 69.1  & 38.1 & 22.0 & 76.4  & 162.5  & 171.9  & 167.2  \\ 
Dynamic-SiC~(ours)                  & 7.1  & 21.3 & 37.4  & 53.5  & 68.0  & 37.5 & 21.7 & 76.0  & 150.1  & 151.7  & 151.0  \\
\hline
\end{tabular}
}
\label{tab:generlization}
\end{table*}

\begin{figure*}[t]
\centering
\includegraphics[width=0.9\linewidth]{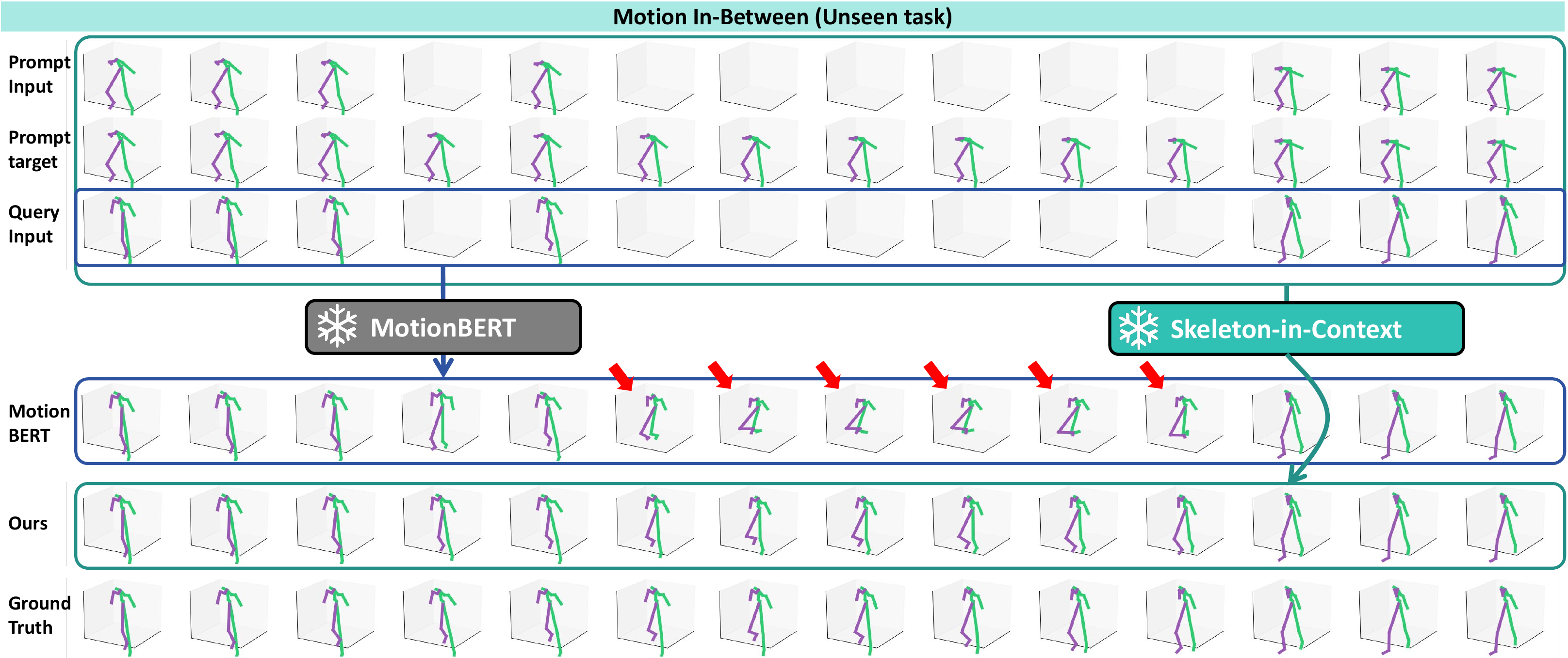}
\centering
\caption{Comparison of generalization. Our SiC completes the missing skeletons according to the customized prompt for an unseen task.
}
\vspace{-1.5em}
\label{fig:generalization}
\end{figure*}

\noindent
\textbf{Datasets and Metrics.}
H3.6M (Human3.6M) is used for the pose estimation and future pose estimation tasks.
AMASS is used for the motion prediction task.
3DPW (3D Pose in the Wild) is used for the joint completion task.
All of the tasks can be evaluated using Mean Per Joint Position Error.

\subsection{Baseline Methods}
In order to make a qualitative comparison with other methods in a relatively fair situation, we reimplement the task-specific SoTA methods in our proposed benchmark. Additionally, we transform recent methods into multi-task models and compare them with our model.

\noindent
\textbf{Copy Method.} 
Following~\cite{bar2022visualprompt,fang2023pic}, we copy the prompt target $T_{i}^{k}$ as prediction results for comparison.

\noindent
\textbf{Multi-Task-Capable Models.}
For task-specific models~\cite{martinez2017simple,xu2023eqmotion,tang2023stcformer} and multi-stage models~\cite{zhu2023motionbert}, they are not able to handle multiple tasks at the same time. Therefore, we re-model them into end-to-end multi-task-capable models to achieve fair comparison.

\noindent
\textbf{In-Context Models.}
Since there are no in-context models excelling in processing multiple human perception tasks, we select the most relative in-context models with our benchmark, Point-In-Context~\cite{fang2023pic} and PointMAE~\cite{pang2022pointmae}, which is proposed in 3D point cloud understanding. Specifically, we take each frame as a token in their framework. For PointMAE, we use the transformed version in ~\cite{fang2023pic}.

\subsection{Compared with SoTA Methods}
\label{sec:mian_results}

\noindent
\textbf{Quantitative Results.}
We report experimental results on four tasks in Tab.~\ref{tab:main_results}. We term Skeleton-in-Context implemented with static/dynamic-TUP as Static/Dynamic-SiC, respectively. For other models, we reproduce them and report the results. As can be seen from Tab.~\ref{tab:main_results}, our proposed model outperforms all multi-task models, achieving state-of-the-art results on the in-context multi-task benchmark. Notably, the results of task-specific models on a single task reflect the difficulty of each task. However, we also exhibit impressive performance compared to task-specific models, demonstrating that our proposed model is an expert in handling skeleton sequence-based multi-tasking.

\noindent
\textbf{Qualitative Results.}
We visualize and compare the results of our SiC and the most recent SoTA model, MotionBERT~\cite{zhu2023motionbert}, which is re-structured as an end-to-end multi-task model for fair comparison. As highlighted in  Fig.~\ref{fig:visualization}, our SiC can generate more accurate poses than MotionBERT according to the provided task-guided prompt.

\subsection{Generalization Capability.}
\label{sec:generalization}

\noindent
\textbf{Generalize to New Datasets.}
To verify the generalization ability of our model, we conduct cross-dataset experiments on a specific task. For instance, our proposed model learns motion prediction on the AMASS dataset under the default benchmark, and then it is used to predict future motion sequences on 3DPW during testing. Similarly, we set up four experiments: motion prediction~(AMASS$\rightarrow$3DPW), pose estimation~(H3.6M$\rightarrow$AMASS, 3DPW), and joint completion~(3DPW$\rightarrow$H3.6M). As shown in Tab.~\ref{tab:generlization}, our model is able to generalize the learned task well to other datasets, while other task-specific models cannot accommodate the large bridge introduced across datasets even if they are specifically trained on this single task. Additionally, benefiting from the dynamic pseudo pose, the Dynamic-SiC exhibits more excellent generalization ability than the Static-SiC.

\noindent
\textbf{Generalize to New Task.}
Furthermore, We construct a prompt that has not appeared in the training set, which performs motion in-between. As Fig.~\ref{fig:generalization} shows, our trained model can complete the missing skeletons according to the customized task-guided prompt. Our SiC generalizes well to a new task, which is unexplored in other works.

\begin{table}[t]
\centering
\setlength\tabcolsep{1mm}
\caption{Effectiveness of the Skeleton Prompt.}
\vspace{-0.5em}
\scalebox{0.76}{
\begin{tabular}{c|cc|cccc}
\hline
\# & TGP & TUP & MP. MPJPE$\downarrow$ & PE. MPJPE$\downarrow$ & JC. MPJPE$\downarrow$ & FPE. MPJPE$\downarrow$ \\ \hline
1 & \XSolidBrush    & \XSolidBrush    &  111.7     & 159.2   & 543.7    & 160.2     \\
2 & \Checkmark    & \XSolidBrush    &  25.0     & 52.5   & 37.7    & 65.2      \\
3 & \XSolidBrush  & \Checkmark   &  26.0     &  52.4  &  37.8   & 78.4      \\ 
\rowcolor{mygray}
4 & \Checkmark    & \Checkmark    &  \textbf{24.4}     & \textbf{51.8}   &   \textbf{36.8}   & \textbf{62.9}       \\ \hline
\end{tabular}
}
\vspace{-1em}
\label{tab:effective_prompt}
\end{table}


\subsection{Ablation Study}

\noindent
\textbf{Effectiveness of the Skeleton Prompt.}
We conduct ablation experiments on our proposed skeleton prompts. As shown in Tab.~\ref{tab:effective_prompt}, the skeleton prompts solve the problem of directly extending previous work to skeleton sequences, which is falling into over-fitting. Meanwhile, applying both TGP and TUP to the model can achieve the best results. Our proposed skeleton prompts replace the random masking training strategy used in previous work. In particular, TUP creates a training environment that does not require the query target, which ensures that the model will not see the ground truth of the query input in advance.

\noindent
\textbf{Prompt Engineering.}
Since previous work has proven that different prompts in context learning affect the performance of the model, we conduct ablation experiments on the choice of prompts. As Tab.~\ref{tab:ablation_TGP} shows, we select the prompt by the feature similarity and MPJPE minimum between the input skeleton sequence and the prompt, respectively. Experimental results show that the random selection method achieves the best results.
Unlike 2D images and 3D point clouds, skeleton-based task data is highly similar, resulting in different prompts playing similar roles in the same task. While the random selection method expands the model's receptive field to the data and achieves better results.

\begin{table}[t]
\centering
\setlength\tabcolsep{1mm}
\caption{Different Prompting~(TGP).}
\vspace{-0.5em}
\scalebox{0.76}{
\begin{tabular}{c|c|cccc}
\hline
\# & TGP & MP. MPJPE$\downarrow$ & PE. MPJPE$\downarrow$ & JC. MPJPE$\downarrow$ & FPE. MPJPE$\downarrow$ \\ \hline
1 & Feature Sim.    &  25.2     & 52.1   & 37.7    & 64.1      \\
2 & Motion Sim.     &  25.2     & 52.1   & 37.8    & 64.1      \\
\rowcolor{mygray}
3 & Random          &  \textbf{24.4}     & \textbf{51.8}   & \textbf{36.8}    & \textbf{62.9}       \\  \hline
\end{tabular}
}
\vspace{-1em}
\label{tab:ablation_TGP}
\end{table}

\noindent
\textbf{Architecture.}
We conduct ablation experiments on the architecture of the model. As shown in Tab.~\ref{tab:ablation_arch}, after exploring different hidden feature dimensions and transformer depths, we find that the model performs best when the feature dimension is $256$, and the depth is $5$.

\begin{table}[t]
\centering
\setlength\tabcolsep{1mm}
\caption{Ablation on model architecture.}
\vspace{-0.5em}
\scalebox{0.76}{
\begin{tabular}{c|c|cccc}
\hline
Arch.                 & Num. & MP. MPJPE$\downarrow$ & PE. MPJPE$\downarrow$ & JC. MPJPE$\downarrow$ & FPE. MPJPE$\downarrow$ \\ \hline
\multirow{4}{*}{Dim.} & 64   & 29.3       & 57.1       & 45.8       & 80.5        \\
                      & 128  & 26.4       & 53.8       & 39.2       & 66.1        \\
                      & \cellcolor{mygray} 256  & \cellcolor{mygray} \textbf{24.4}       & \cellcolor{mygray} 51.8       & \cellcolor{mygray} \textbf{36.8}       & \cellcolor{mygray} \textbf{62.9 }       \\
                      & 512  & 24.7       & \textbf{50.7}       & 37.6       & 63.0        \\ \hline
\multirow{4}{*}{Dep.} & 3    & 25.1       & 53.7       & 39.2       & 65.6        \\
                      & 4    & 25.1       & 52.2       & 38.6       & 64.7        \\
                      & \cellcolor{mygray} 5    & \cellcolor{mygray} \textbf{24.4}       & \cellcolor{mygray} \textbf{51.8}       & \cellcolor{mygray} \textbf{36.8}      & \cellcolor{mygray} \textbf{62.9}        \\
                      & 6    & 24.7       & 52.9       & 38.1       & 63.7        \\ \hline
\end{tabular}
}
\vspace{-0.5em}
\label{tab:ablation_arch}
\end{table}


\subsection{In-Context Initial Pose}
\begin{figure}[t]
\centering
\includegraphics[width=0.99\linewidth]{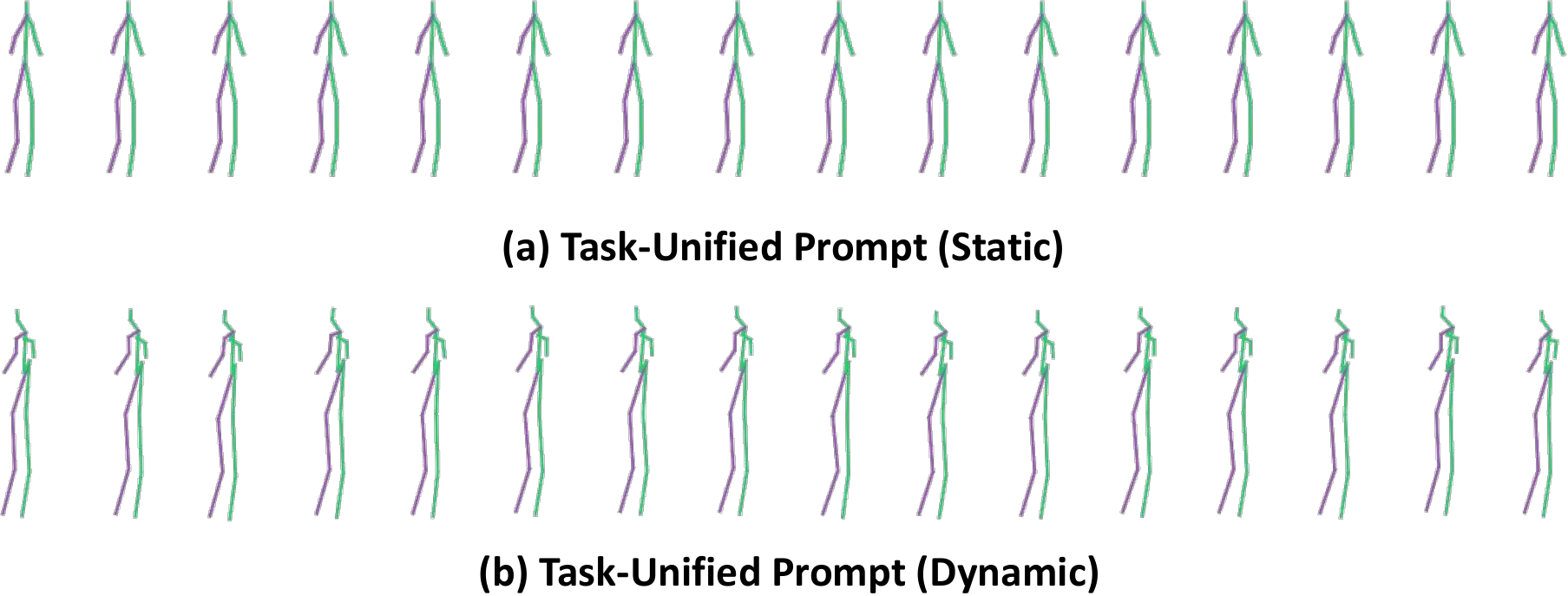}
\centering
\vspace{-0.5em}
\caption{(a) The visualization of Static-TUP. (b) The concrete visual skeleton sequence of Dynamic-TUP.}
\vspace{-1.5em}
\label{fig:TUP_Pose}
\end{figure}

We conduct an in-depth exploration of Dynamic-TUP. We extract the trained Dynamic-TUP weights, then randomly initialize a $(F, J, 3)$ tensor and use the trained encoder to increase its feature dimension to C-dimension. Finally, an optimizer is used to optimize the tensor and minimize its feature similarity with the trained Dynamic-TUP. The top of Fig.~\ref{fig:TUP_Pose} represents the visualization of Static-TUP, while the bottom is the Dynamic-TUP. The concrete skeleton of dynamic-TUP is similar to the mean pose Static-TUP of the training set, which demonstrates that the Dynamic-TUP learned by our model is actually the prior knowledge whose receptive field is all ground truth in the training set. This prior knowledge helps our model learn more smoothly on multiple tasks, and adding learnable modules helps our model generalize to other datasets or new tasks~(Sec.~\ref{sec:generalization}).

\section{Conclusion}
\label{sec:conclusion}

In this work, we propose the Skeleton-in-Context, designed to process multiple skeleton-base tasks simultaneously after just one training time. 
Specifically, we build a novel skeleton-based in-context benchmark covering various tasks. In particular, we propose skeleton prompts composed of TGP and TUP, which solve the overfitting problem of skeleton sequence data trained under the training framework commonly applied in previous 2D and 3D in-context models. 
Besides, we demonstrate that our model can generalize to different datasets and new tasks, such as motion completion. 
We hope our research builds the first step in the exploration of in-context learning for skeleton-based sequences, which paves the way for further research in this area.

\noindent
\textbf{Acknowledgement}
This work is supported by National Natural Science Foundation of China (No. 62203476), Natural Science Foundation of Shenzhen (No. JCYJ20230807120801002). This project is also supported by the interdisciplinary doctoral grants (iDoc 2021-360) from the Personalized Health and Related Technologies (PHRT) of the ETH domain, Switzerland.

\normalem
{
    \small
    \bibliographystyle{ieeenat_fullname}
    \bibliography{main}
}

\clearpage
\appendix

\begin{strip}
\begin{table}[H]
\hsize = \textwidth
\small
\centering
\vspace{-3em}
\caption{Importance of our proposed task-guided prompt (TGP) and task-unified prompt (TUP) employed in Skeleton-in-Context (SiC). TGP and TUP are important designs as they address the challenge of multi-task training. Please refer to Sec.~\ref{sec:supp_add_exp} for more discussion.
%
}
\vspace{-0.5em}
\setlength\tabcolsep{1mm}
\begin{tabular}{lcccccccccccccc}
\hline
\multicolumn{1}{l|}{\multirow{3}{*}{Methods}}  & \multicolumn{6}{c|}{MP. (AMASS)}     & \multicolumn{2}{c|}{PE. (H3.6M)} & \multicolumn{3}{c|}{JC. (3DPW)} & \multicolumn{3}{c}{FPE. (H3.6M)} \\
\multicolumn{1}{c|}{}   & \multicolumn{6}{c|}{MPJPE $\downarrow$}     & MPJPE$\downarrow$ & \multicolumn{1}{c|}{N-MPJPE$\downarrow$}   & \multicolumn{3}{c|}{MPJPE $\downarrow$} & \multicolumn{3}{c}{MPJPE $\downarrow$} \\
\multicolumn{1}{c|}{} & 80ms & 160ms & 200ms & 320ms & 400ms & \multicolumn{1}{c|}{Avg.} & Avg. & \multicolumn{1}{c|}{Avg.}        & 40\%   & 60\%   & \multicolumn{1}{c|}{Avg.}   & 200ms  & 300ms & Avg.      \\ \hline
\multicolumn{15}{c}{Task-Specific Model: one architecture for one task}   \\ \hline
\rowcolor{mygray}
\multicolumn{1}{l|}{SiC~(w/o TGP\&TUP)} & 57.7 & 46.2 & 41.6 & 36.7 & 47.7 & \multicolumn{1}{c|}{46.0} & 56.5 & \multicolumn{1}{c|}{45.4} & 49.7 & 59.8 & \multicolumn{1}{c|}{54.7} & 62.5 & 72.6 & 67.6 \\ \hline
\multicolumn{15}{c}{Multi-Task Model: one architecture (w. multiple task-specific heads) for multiple tasks}    \\ \hline
\multicolumn{1}{l|}{SiC~(w/o TGP\&TUP)} & 62.9 & 49.8 & 44.3 & 36.3 & 46.5 & \multicolumn{1}{c|}{47.9} & 59.1 & \multicolumn{1}{c|}{47.4} & 51.3 & 61.6 & \multicolumn{1}{c|}{56.5} & 76.4 & 86.8 & 81.6 \\ \hline
\multicolumn{15}{c}{In-Context Model: one task-agnostic architecture for multiple tasks}    \\ \hline
\multicolumn{1}{l|}{Static-SiC}       & \textbf{\textcolor{blue}{11.0}}   & 19.0     & 22.7   & 32.6 & 37.8 & \multicolumn{1}{c|}{24.6}   & \textbf{\textcolor{blue}{51.6}}  & \multicolumn{1}{c|}{\textbf{\textcolor{blue}{41.8}}}     & 29.8      & 44.7      & \multicolumn{1}{c|}{37.2}      & 59.8    & 67.2    & 63.5           \\ 
\multicolumn{1}{l|}{Dynamic-SiC}     & \textbf{\textcolor{blue}{11.0}}   & \textbf{\textcolor{blue}{18.9}}     & \textbf{\textcolor{blue}{22.6}}   & \textbf{\textcolor{blue}{32.2}} & \textbf{\textcolor{blue}{37.4}} & \multicolumn{1}{c|}{\textbf{\textcolor{blue}{24.4}}}   & 51.8  & \multicolumn{1}{c|}{42.5}     & \textbf{\textcolor{blue}{29.5}}      & \textbf{\textcolor{blue}{44.0}}      & \multicolumn{1}{c|}{\textbf{\textcolor{blue}{36.8}}}      & \textbf{\textcolor{blue}{59.2}}    & \textbf{\textcolor{blue}{66.5}}    & \textbf{\textcolor{blue}{62.9}}       \\ 
\hline
\end{tabular}
\vspace{-1.5em}
\label{tab:supp_results}
\end{table}
\end{strip}


\noindent
\textbf{Overview.}
The supplementary material includes:

\begin{itemize}
    \item Section~\ref{sec:supp_exp_details} presents more details on our experiments and a more detailed description of the datasets.
    \item Section~\ref{sec:supp_add_exp} verifies the effectiveness of our Skeleton-in-Context on multi-task synergistic training.
    \item Section~\ref{sec:supp_analysis} provides more detailed results both quantitatively and qualitatively.
\end{itemize}

\section{Experimental Details}
\label{sec:supp_exp_details}

\subsection{Implementation Details}
We implement the proposed Skeleton-in-Context model with the number of layers $N$ = 5, number of attention heads $H$ = 8, and hidden feature dimension $C$ = 256. For each prompt input/target and query input/target, the sequence length is $F$ = 16.
We implement Skeleton-in-Context with PyTorch. In the default setting, during both training and evaluation, for each query pair, we randomly select a prompt pair from the training set of the same task as the query pair. We use an AdamW optimizer with a linearly decaying learning rate, starting at 0.0002 and decreasing by 1\% after every epoch.
All tasks are unified into a one-off, end-to-end training process without any task-specific designs. The training takes about 6 hours for 120 epochs on 4 NVIDIA GeForce RTX 4090 GPUs.

\subsection{Datasets and Metrics}

\noindent
\textbf{H3.6M (Human3.6M)} is used for the pose estimation and future pose estimation tasks. It is a large-scale dataset that contains 3.6 million video frames of actions involving 15 types and 7 actors. Following previous works~\cite{zhu2023motionbert}, we use subjects 1, 5, 6, 7, and 8 for training and subjects 9 and 11 for testing and preprocessing the poses. After preprocessing, each pose has 17 joints.
We use the Stacked Hourglass (SH) networks~\cite{newell2016stacked} to extract the 2D skeletons from videos as input. For pose estimation, we expect the model to estimate the corresponding 3D skeletons. For future pose estimation, we expect the model to predict and estimate at the same time the future 3D skeletons in a time range of 300ms, given the history 300ms of 2D skeletons.
As we use the same dataset on two tasks, multi-tasking is harder as the model needs further context to correctly perceive and then accomplish the task. In Skeleton-in-Context we use prompts to guide the model to learn in context.

\noindent
\textbf{AMASS} is used for the motion prediction task. It integrates most of the existing marker-based Mocap datasets, which are parameterized with a unified representation. We follow common practices in human motion prediction~\cite{mao2020history,guo2023back} to use AMASS-BMLrub for testing and the rest of the datasets for training.
For motion prediction, we expect the model to predict the future 3D motion sequence, given the 3D history motion sequences. The time ranges of the future and history motion sequences are both 400ms, which is 10 frames under the frame rate of 25 fps. As the model requires the sequence to be of 16 frames, we pad the last poses 6 times. 

\noindent
\textbf{3DPW (3D Pose in the Wild)} is used for the joint completion task. It has more than 51k frames with 3D annotations for challenging indoor and outdoor activities.
For joint completion, we construct 2 settings, where we randomly mask 40\% and 60\% of all the joints. We expect the model to reconstruct the missing joints.

\noindent
\textbf{Metrics.}
For Motion Prediction~(MP.), Joint Completion~(JC.), and Future Pose Estimation~(FPE.), we report Mean Per Joint Position Error~(mm)~\cite{ionescu2013h36m}. For Pose Estimation~(PE.), we additionally report another indicator, Normalized Mean Per Joint Position Error~(N-MPJPE)~\cite{zhu2023motionbert}.

\begin{table*}[t]
\small
\centering
\setlength\tabcolsep{1mm}
\caption{Detailed results of pose estimation between our Skeleton-in-Context and multi-task models re-structured from task-specific models or multi-stage models. We report Mean Per Joint Position Error (MPJPE).}
\scalebox{0.87}{
\begin{tabular}{l|c|ccccccccccccccc|c}
\hline
Methods     & Venues  & Dire. & Disc. & Eat. & Greet & Phone & Photo & Pose & Purch. & Sit. & SitD. & Smoke & Wait & Walk  & WalkD. & WalkT. & Avg  \\ \hline
siMLPe~\cite{martinez2017simple}      & WACV'23 & 56.4  & 65.3  & 69.2  & 68.7  & 71.7  & 98.3  & 57.9  & 66.8  & 91.7  & 129.9 & 68.4  & 72.2  & 61.3  & 77.1  & 63.5  & 74.6 \\
EqMotion~\cite{xu2023eqmotion}    & CVPR'23 & 141.4 & 151.9 & 175.1 & 154.0 & 165.2 & 173.4 & 131.0 & 194.1 & 201.7 & 219.0 & 159.6 & 151.6 & 129.3 & 160.9 & 141.8 & 163.3 \\
STCFormer~\cite{tang2023stcformer}   & CVPR'23 & 46.3  & 52.7  & 51.2  & 48.5  & 53.7  & 70.3  & 48.3  & 48.3  & 72.9  & 100.0 & 54.6  & 50.3  & 38.4  & 56.7  & 41.8  & 55.6 \\
GLA-GCN~\cite{yu2023glagcn}     & ICCV'23 & 58.1  & 71.4  & 73.0  & 67.7  & 78.9  & 97.3  & 60.9  & 73.2  & 90.1  & 125.4 & 75.0  & 73.3  & 74.0  & 84.3  & 76.3  & 78.6 \\
MotionBERT~\cite{zhu2023motionbert}  & ICCV'23 & 47.3  & 52.7  & 49.8  & 49.0  & 56.9  & 73.2  & 48.7  & 49.5  & 67.8  & 94.4  & 55.3  & 52.7  & 40.2  & 57.2  & 42.8  & 55.8 \\ \hline
Static-SiC~(ours)  & -       & \textbf{\textcolor{blue}{45.8}}  & 52.9  & \textbf{\textcolor{blue}{46.6}}  & \textbf{\textcolor{blue}{46.6}}  & 54.6  & \textbf{\textcolor{blue}{66.8}}  & 47.9  & 47.0  & \textbf{\textcolor{blue}{60.4}}  & \textbf{\textcolor{blue}{74.5}}  & \textbf{\textcolor{blue}{51.7}}  & 49.9  & \textbf{\textcolor{blue}{37.0}}  & \textbf{\textcolor{blue}{51.9}}  & \textbf{\textcolor{blue}{40.0}}  & \textbf{\textcolor{blue}{51.6}} \\
Dynamic-SiC~(ours) & -       & 47.0  & \textbf{\textcolor{blue}{52.1}}  & 48.8  & 47.4  & \textbf{\textcolor{blue}{52.6}}  & 69.1  & \textbf{\textcolor{blue}{47.1}}  & \textbf{\textcolor{blue}{46.6}}  & \textbf{\textcolor{blue}{60.4}}  & \textbf{\textcolor{blue}{74.5}}  & 52.3  & \textbf{\textcolor{blue}{49.1}}  & 37.5  & 52.1  & 40.1  & 51.8 \\ \hline
\end{tabular}
}
\label{tab:detail_PE}
\end{table*}
\begin{table*}[t]
\small
\centering
\setlength\tabcolsep{1mm}
\caption{Detailed results of future pose estimation between our Skeleton-in-Context and multi-task models re-structured from task-specific models or multi-stage models. We report Mean Per Joint Position Error (MPJPE).}
\scalebox{0.9}{
\begin{tabular}{l|c|ccccccccccccccc|c}
\hline
Methods     & Venues  & Dire. & Disc. & Eat. & Greet & Phone & Photo & Pose & Purch. & Sit. & SitD. & Smoke & Wait & Walk  & WalkD. & WalkT. & Avg  \\ \hline
siMLPe~\cite{martinez2017simple}      & WACV'23 & 61.2  & 73.4  & 63.9 & 67.6  & 64.2  & 79.9  & 62.2 & 76.1   & 68.8 & 88.0  & 62.1  & 65.1 & 69.7  & 80.4   & 71.9   & 66.4 \\
EqMotion~\cite{xu2023eqmotion}    & CVPR'23 & 73.5  & 82.8  & 92.3 & 85.8  & 86.9  & 94.8  & 70.5 & 100.7  & 94.1 & 102.1 & 82.8  & 81.0 & 104.5 & 93.1   & 101.1  & 88.6 \\
STCFormer~\cite{tang2023stcformer}   & CVPR'23 & 59.8  & 68.9  & 58.3 & 69.4  & 62.7  & 80.2  & 63.5 & 85.2   & 72.0 & 90.4  & 62.1  & 68.1 & 72.4  & 81.5   & 80.0   & 67.0 \\
GLA-GCN~\cite{yu2023glagcn}     & ICCV'23 & 59.9  & 68.4  & 67.0 & 70.6  & 67.6  & 82.6  & 67.7 & 70.5   & 62.9 & 79.8  & 64.4  & 66.6 & 73.1  & 80.2   & 86.5   & 64.4 \\
MotionBERT~\cite{zhu2023motionbert}& ICCV'23 & 63.2  & 74.9  & 66.9 & 105.7 & 70.2  & 93.3  & 94.4 & 93.3   & 67.8 & 92.0  & 68.7  & 89.8 & 83.6  & 129.0  & 78.3   & 86.2 \\ \hline
Static-SiC~(ours)  & -       & \textbf{\textcolor{blue}{59.3}}  & 68.4  & \textbf{\textcolor{blue}{55.1}} & \textbf{\textcolor{blue}{67.5}}  & 60.3  & 79.8  & 61.3 & 69.9   & 62.8 & \textbf{\textcolor{blue}{78.2}}  & \textbf{\textcolor{blue}{59.3}}  & 60.6 & 48.9  & 71.3   & 51.0   & 63.5 \\
Dynamic-SiC~(ours) & -       & 59.7  & \textbf{\textcolor{blue}{67.3}}  & 55.4 & 67.8  & \textbf{\textcolor{blue}{57.8}}  & \textbf{\textcolor{blue}{79.1}}  & \textbf{\textcolor{blue}{59.4}} & \textbf{\textcolor{blue}{68.6}}   & \textbf{\textcolor{blue}{62.6}} & 79.4  & 59.4  & \textbf{\textcolor{blue}{60.4}} & \textbf{\textcolor{blue}{48.4}}  & \textbf{\textcolor{blue}{70.4}}   & \textbf{\textcolor{blue}{49.9}}   & \textbf{\textcolor{blue}{62.9}} \\ \hline
\end{tabular}
}
\label{tab:detail_FPE}
\end{table*}

\section{Multi-Task Synergistic Training}
\label{sec:supp_add_exp}
In this section, we address the challenge brought by multi-task training and demonstrate the effectiveness of our model under multi-task learning of human motion representations. First, we analyze the effect of negative transfer and how it can limit multi-task training. Next, we evaluate the effectiveness of our proposed task-guided and task-unified prompts (TGPs and TUPs) in addressing this challenge and facilitating collaborative training between multiple tasks.

\subsection{Negative Transfer in Multi-Task}
A straight way to train a generalist multi-task model is to collect the data from multiple tasks, format them into the same shape, and directly train the model on them. However, as explained in \cite{wu2023ppt}, a phenomenon named \textit{negative transfer} in multi-dataset training will occur and lead to poor results in the above training method. As different tasks have their own unique goals, they may confuse the model during training without guidance from context, leading to a performance drop in testing. To demonstrate the effect of negative transfer, as shown in Tab.~\ref{tab:supp_results}, we train and test the backbone of our SiC separately on four tasks, whose results are in Tab.~\ref{tab:supp_results} with gray background. Note that the backbone here does not include TUP and TGP. When we train the four tasks together, the performance of the model decreases due to the data variability between the individual tasks, which can be attributed to the \textit{negative transfer} phenomenon~\cite{wu2023ppt}. The experiments show that, without context guidance from prompts (TUP and TGP), the multi-task training is limited by negative transfer and not able to achieve satisfactory performance.

\subsection{Multi-Task Synergistic Training}
Our Skeleton-in-Context enables multiple tasks to be trained in a unified way, which we term Multi-Task Synergistic Training. With our proposed Task-Guided Prompt~(TGP), SiC can perceive context (the corresponding task-specific information) from TGP, and accomplish the task accurately. Meanwhile, in order to deal with data from different datasets and tasks, which have different patterns and may affect the performance if not processed properly, we introduce the Task-Unified Prompt~(TUP) as prior knowledge of the query target. As a task-agnostic module, the TGP is able to encode the unified human motion representations of various tasks. As Tab.~\ref{tab:supp_results} shows, with the proposed TGP and TUP, our SiC is able to achieve impressive results on four tasks simultaneously.

\section{Detailed Results and Visualization}
\label{sec:supp_analysis}

\subsection{Detailed Results}
We report the action-wise results in pose estimation and future pose estimation on H3.6M~\cite{ionescu2013h36m}. As shown in Tab.~\ref{tab:detail_PE} and Tab.~\ref{tab:detail_FPE}, our model outperforms other multi-task models in each action on H3.6M, which verifies our model's ability to handle multi-tasks. Additionally, Dynamic-SiC performs better than Static-SiC in most actions.

\subsection{More Visualization}
\begin{figure*}[t]
\centering
\includegraphics[width=0.93\linewidth]{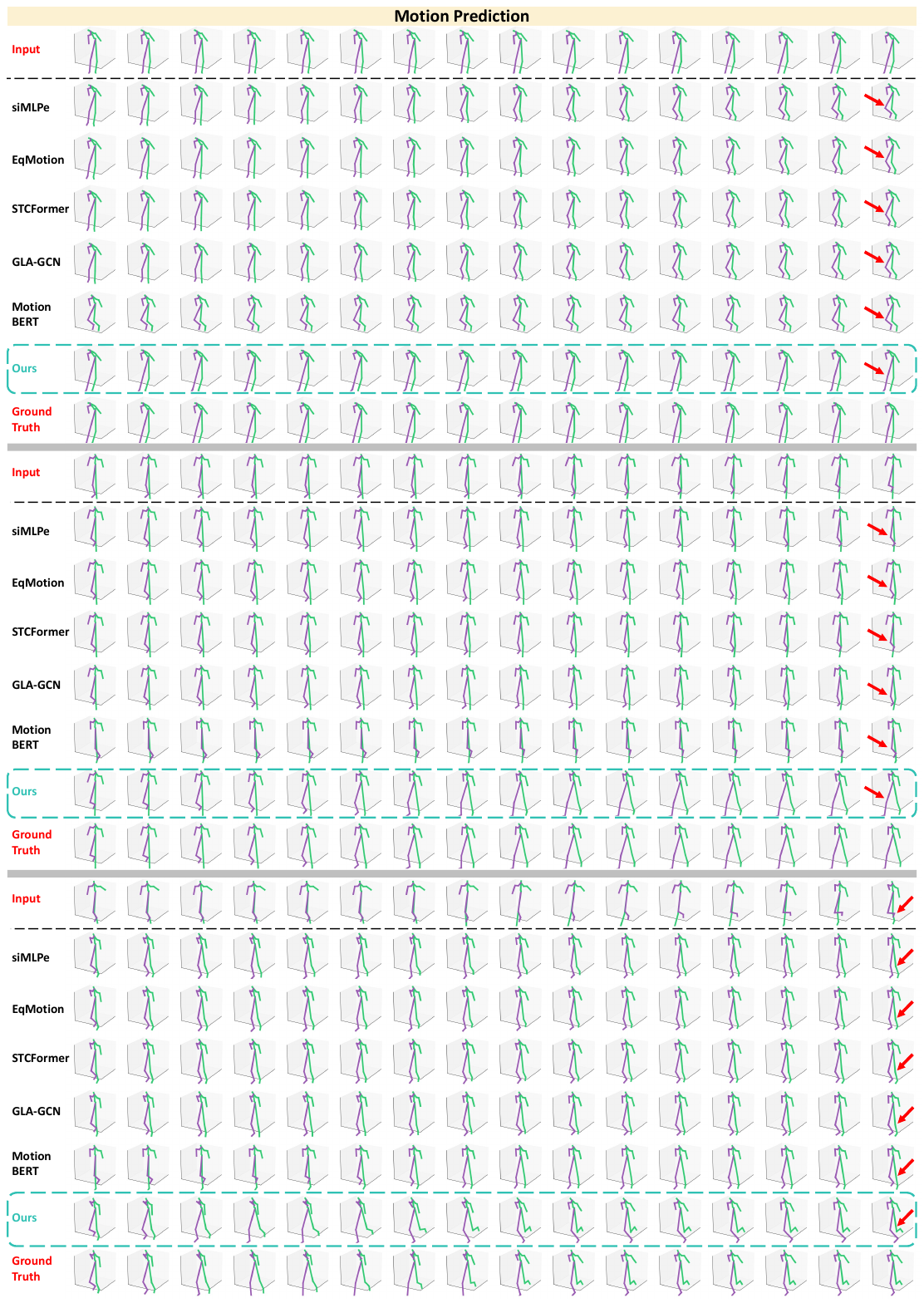}
\centering
\vspace{-1em}
\caption{Comparison of visualization between multi-task models and our Skeleton-in-Context on motion prediction.
}
\vspace{-1.5em}
\label{fig:supp_mp}
\end{figure*}

\begin{figure*}[t]
\centering
\includegraphics[width=0.93\linewidth]{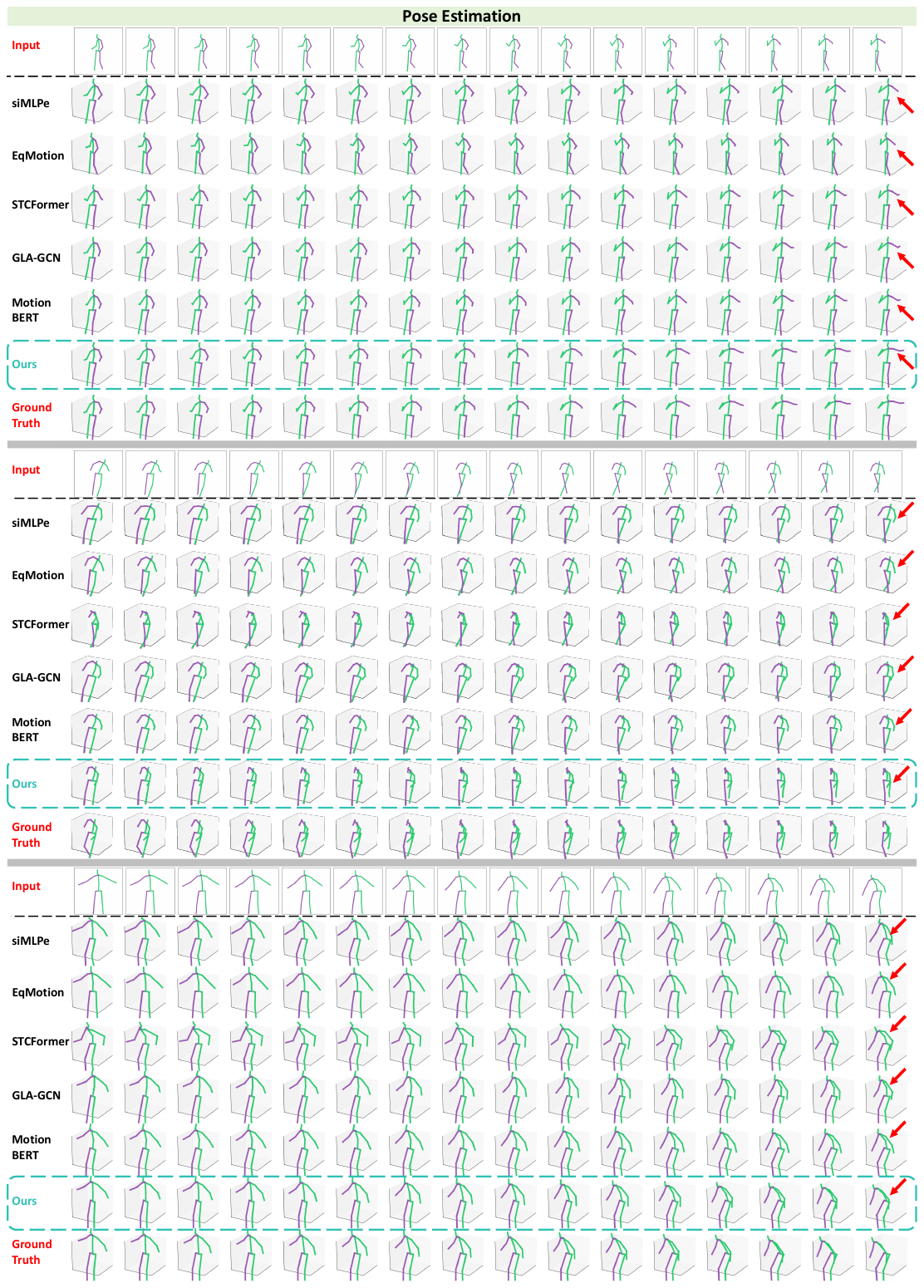}
\centering
\vspace{-1em}
\caption{Comparison of visualization between multi-task models and our Skeleton-in-Context on pose estimation.
}
\vspace{-1.5em}
\label{fig:supp_pe}
\end{figure*}

\begin{figure*}[t]
\centering
\includegraphics[width=0.93\linewidth]{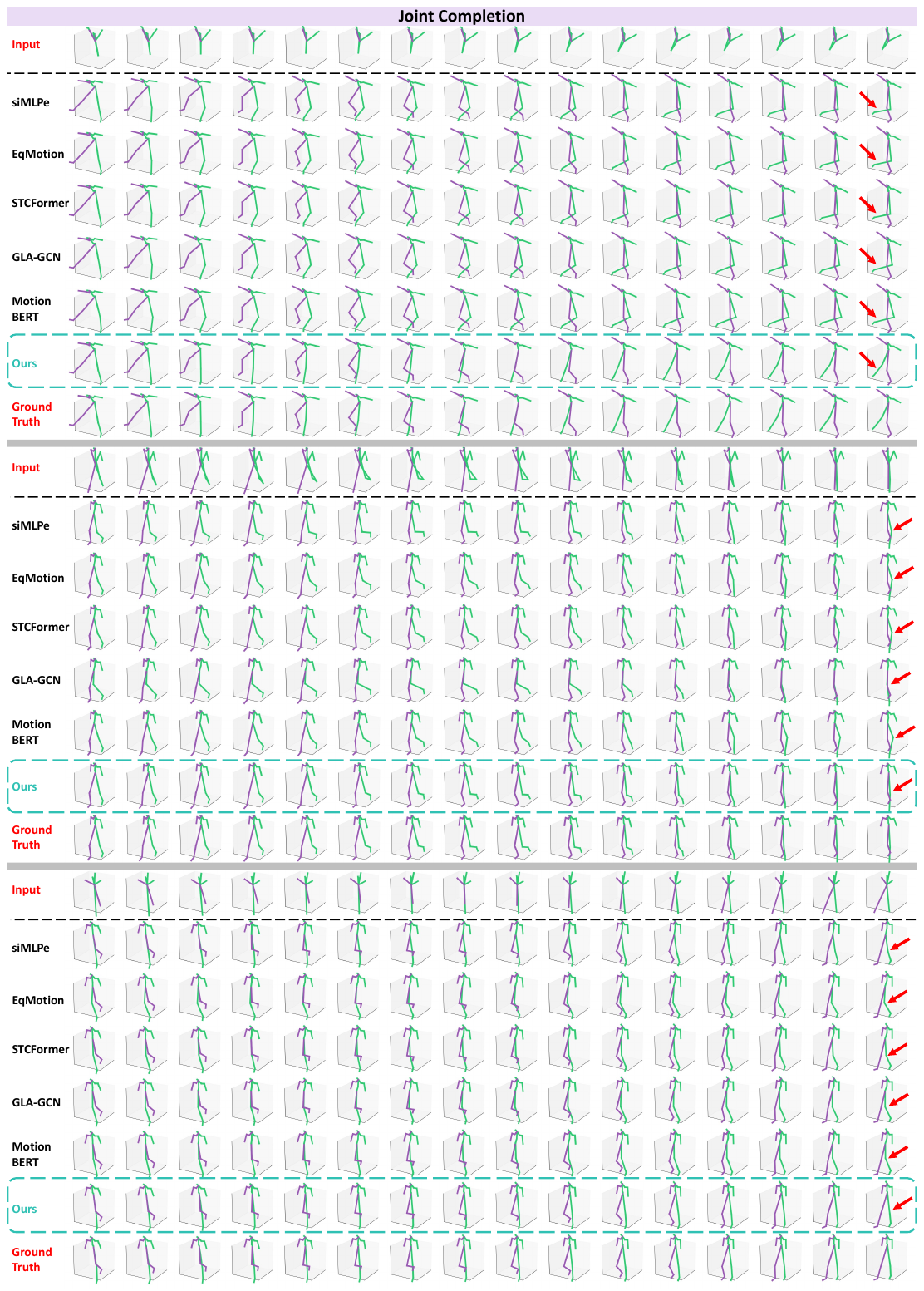}
\centering
\vspace{-1em}
\caption{Comparison of visualization between multi-task models and our Skeleton-in-Context on joint completion.
}
\vspace{-1.5em}
\label{fig:supp_jc}
\end{figure*}

\begin{figure*}[t]
\centering
\includegraphics[width=0.93\linewidth]{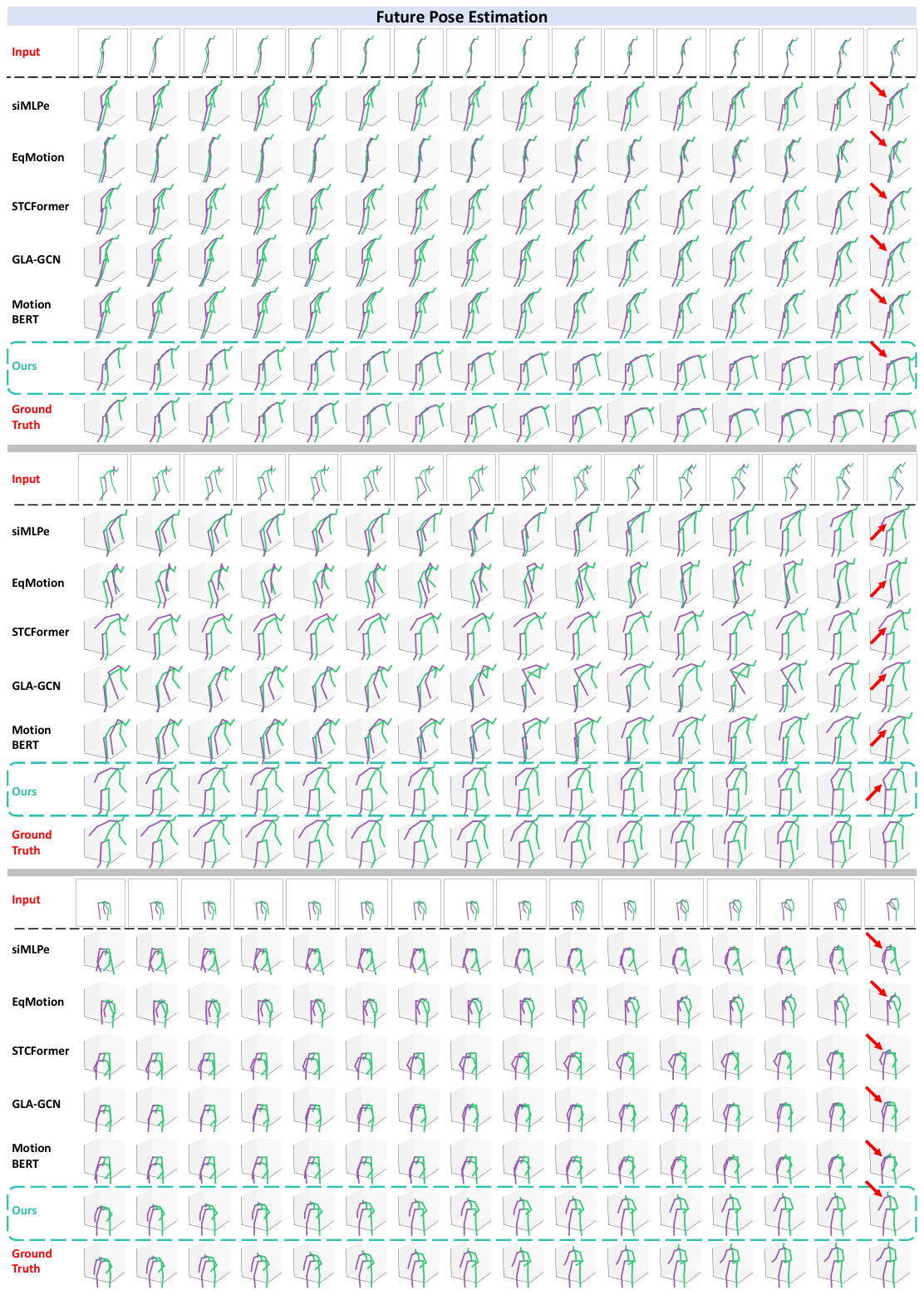}
\centering
\vspace{-1em}
\caption{Comparison of visualization between multi-task models and our Skeleton-in-Context on future pose estimation.
}
\vspace{-1.5em}
\label{fig:supp_fpe}
\end{figure*}

We provide more visualization results on four tasks in comparison with all the multi-task baselines that we mention in the main text, as shown in Fig.~\ref{fig:supp_mp}, \ref{fig:supp_pe}, \ref{fig:supp_jc}, \ref{fig:supp_fpe}. 
Other models are obviously unable to balance the resources of each task during training, which leads to unsatisfactory results.

\end{document}